\documentclass[letterpaper, 10 pt, conference]{ieeeconf}  
\IEEEoverridecommandlockouts

\usepackage{graphics} 
\usepackage{epsfig} 
\usepackage{mathptmx} 
\usepackage{times} 
\usepackage{amsmath} 
\usepackage{amssymb}  
\usepackage{caption}
\usepackage{subcaption}
\usepackage{xcolor}
\usepackage{cite}
\usepackage{float}
\usepackage{soul}
\usepackage{multirow}

\newcommand{\argmin}[1]{\underset{#1}{\operatorname{arg}\,\operatorname{min}}\;}

\usepackage{tikz,xcolor,hyperref}
\definecolor{lime}{HTML}{A6CE39}
\DeclareRobustCommand{\orcidicon}{
	\begin{tikzpicture}
	\draw[lime, fill=lime] (0,0) 
	circle [radius=0.16] 
	node[white] {{\fontfamily{qag}\selectfont \tiny ID}};
	\draw[white, fill=white] (-0.0625,0.095) 
	circle [radius=0.007];
	\end{tikzpicture}
	\hspace{-2mm}
}
\foreach \x in {A, ..., Z}{\expandafter\xdef\csname orcid\x\endcsname{\noexpand\href{https://orcid.org/\csname orcidauthor\x\endcsname}
			{\noexpand\orcidicon}}
}

\title{\LARGE \bf{Autonomous Vehicle Calibration via Linear Optimization}}

\author{Georg Novotny$^{1, 2}$\orcidB \emph{Student Member, IEEE}, Yuzhou Liu$^{1}$\orcidF \emph{Student Member, IEEE},\\
Wilfried W\"ober$^{2,3}$\orcidD \emph{Student Member, IEEE},
and Cristina Olaverri-Monreal$^{1}$\orcidE{} \emph{Senior Member, IEEE}%
\thanks{$^1$ Johannes Kepler University Linz; Chair Sustainable Transport Logistics 4.0, Altenberger Straße 69, 4040 Linz, Austria.
	\texttt{\{georg.novotny, cristina.olaverri-monreal\}@jku.at}}%
\thanks{$^2$ UAS Technikum Wien, H\"ochstaedtplatz 6, 1200  Vienna, Austria
	\texttt{\{georg.novotny, wilfried.woeber\}}@technikum-wien.at}
\thanks{$^3$ Department of Integrative Biology and Biodiversity Research, Institute of Integrative Conservation Research, University of Natural Resources and Life Sciences, Gregor Mendel Str. 33, 1080 Vienna, Austria
    \texttt{\{wilfried.woeber@technikum-wien.at\}}}
}

\begin{document}

\maketitle
\thispagestyle{empty}
\pagestyle{empty}

\begin{abstract}
In navigation activities, kinematic parameters of a mobile vehicle play a significant role. Odometry is most commonly used for dead reckoning. However, the unrestricted accumulation of errors is a disadvantage using this method. As a result, it is necessary to calibrate odometry parameters to minimize the error accumulation. This paper presents a pipeline based on sequential least square programming to minimize the relative position displacement of an arbitrary landmark in consecutive time steps of a kinematic vehicle model by calibrating the parameters of applied model. Results showed that the developed pipeline produced accurate results with small datasets.

\end{abstract}

\section{INTRODUCTION}\label{sec:1}
The calculation of motion is one of the most fundamental and challenging tasks for intelligent vehicles (IV) \cite{olaverri2017road}.
In order to create safe and reliable autonomous behavior of vehicles the motion estimation must be as accurate as possible.
In the most fundamental manifestation the vehicle motion is derived by the integration of wheel motion relying on parameters such as wheel radius and baseline \cite{Siegwart:AMR}.
This results in ego motion estimation (dead reckoning) which is needed for autonomous navigation \cite{OdomNav}, mapping \cite{OdomSlam1, OdomSlam2} and obstacle avoidance \cite{Obstacle1, Obstacle2} for intelligent vehicles \cite{8718624}.
In addition, the mobile robotic community proposes methodologies for motion estimation based on probability theory known as probabilistic robotics\cite{Thrun:PR}.
Based on those models, machine learning is used to obtain motion from sensor data \cite{Wober2020AutonomousKinematics,Ko:GPPF}.
Nevertheless, due to limited explainablitiy of recent machine learning models \cite{Samek:XAI1,Samek:XAI3,Montavon:XAI1} classic models relying on deterministic kinematic properties are still used in IVs \cite{KinBicycle}.

However, the major drawback of these systems 
is the error accumulation over time which is primarily driven by systematic errors due to production inaccuracies, abrasion or inaccurate computer-aided design (CAD) data \cite{lee2014accurate, Sousa2020, 1512356, UMBmark, antonelli2007deterministic}.
The authors of the study in \cite{UMBmark} argued, that vehicle calibration can be used to limit the aforementioned source of error.
However, calibrating vehicles is a tedious task and must be done in a pre-defined track or on single arbitrary paths \cite{Sousa2020}.

We show in the presented paper that it is possible to calibrate a vehicle relying on internal and external sensor data, and we contribute to the state of the art of parameter calibration by providing a novel pipeline for the optimization of vehicle parameters such as wheel radii or baselines in order to (re)calibrate the motion estimation function of intelligent vehicles. 
Our proposed framework uses sensor data to reduce potential drift by observing the environment and internal sensor data. 
For this reduction we use wheel encoder data, an initial guess of vehicle parameters and a distance measurement to an arbitrary landmark. Further the proposed approach does not rely on a predefined trajectory and end-point optimization but can be applied in any area where one arbitrary landmark can be tracked during the data collection. Finally, it uses a general landmark description. To this end we rely on range sensor data and euclidean clustering to track the landmark over time. The framework can also be extended using deep learning alternatives e.g. \cite{DLTracking}.

\begin{figure}[!t]
	\centering
	\includegraphics[width=0.48\textwidth]{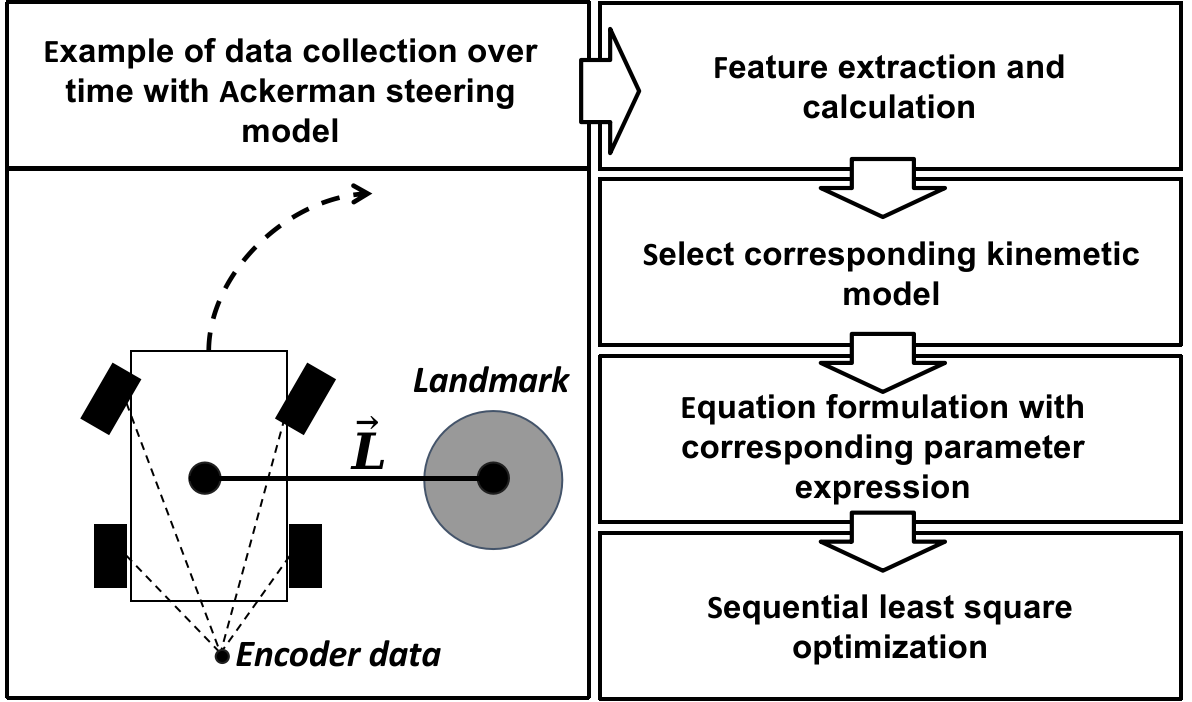}
	\caption{Visualization of the proposed pipeline. 
	We calibrate the vehicles kinematic parameter relying on a detected landmark as well as encoder sensors.}
	\label{fig:pipeline}
\end{figure}

The proposed pipeline is based on the change of the landmark's pose measured by the vehicle's sensor system and the data of the wheel encoders.
These measurements were used to optimize the kinematic parameters of a vehicle.
The proposed pipeline is visualized in Fig. \ref{fig:pipeline}.
To this end we applied linear optimization \cite{Optimizer} in order to estimate vehicle's kinematic parameters relying on the rotational velocity of actuators such as wheels and/or steering angles as well as on the relative pose of an arbitrary landmark. 
Our pipeline was tested with two kinematic configurations of autonomous vehicles.
For this purpose we used simulated data.

This paper is structured as follows: In the following section we describe related literature in the field of parameter calibration for vehicles.
In section \ref{sec:3} we describe the proposed approach followed by experimental setup in section \ref{sec:4}. 
Section \ref{sec:5} presents the results and section \ref{sec:6} concludes the paper outlining future research.
\section{Related Literature}\label{sec:2}
The manual process of kinematic parameter calibration for vehicles is a time consuming work \cite{galasso2019efficient} thus automatic calibration of autonomous vehicles is an active research area.
Classic methods for parameter calibration \cite{Sousa2020} have several limitations in the sense that they  are either defined by a given path with certain
motion sequences (e.g. \cite{LEE2010582}) or have to track the actual position of the vehicle with additional sensors (e.g. \cite{robotics5040023, Martinelli2003}).

In \cite{LEE2010582} a calibration method of the wheel radii and track distance of car-like mobile robots (CLMR) was proposed that reduces systematic errors of dead reckoning by calibrating the kinematic parameters via the average final pose displacement of a predefined track.

\cite{galasso2019efficient} provided an intrinsic and extrinsic calibration method for automated guided vehicles with four wheels that follow dual drive kinematics or Ackerman drive. 
Their method computes the expected trajectories based on the state of the wheels and the change of heading which was observed by an on-board range-sensor using a loop closure based registration method. Both intrinsic and extrinsic calibration parameters were obtained in a calibration process of about 15 minutes, by closed-form solutions of least-square optimization using the model equations derived for the particular kinematics and the loop closure based pose updates.

In \cite{kummerle2012simultaneous} a calibration approach was presented that estimates the wheel radii, sensor positions as well as the robot's position online during Simultaneous Localization and Mapping. The authors achieved this by relying on a probabilistic approach 
which allows for online radii estimation even when the load of the robot or its ground surface changes. The key to their method is to utilize the map estimate as a calibration pattern, then apply a least-squares algorithm to constantly adjust the map, trajectory, and robot parameter estimates.

\cite{Deray2019JointManifold} presented an approach that calibrates, for a differential drive vehicle, the extrinsic parameters of an exteroceptive sensor, which is capable of sensing ego-motion, as well as the intrinsic parameters of its odometry motion model simultaneously. The core idea was to use the principle of recursive pre-integration theory in combination with Lie theory enabling a true on-manifold estimation of the parameters.

We contribute to the state of the art in autonomous vehicle calibration by applying linear optimization in order to estimate a vehicle's kinematic parameters by relying on the rotational velocity of wheels as well as on the relative pose of an arbitrary landmark. This approach constitutes an advancement in the field because it only utilizes 40 seconds of data for a simulated differential drive vehicle, and 90 seconds for a simulated Ackerman vehicle. Further the proposed approach does not rely on a fixed trajectory and end-point optimization but it can be applied in any area where one arbitrary landmark can be
tracked during the data collection.

\section{Methods}\label{sec:3}
We interpreted the vehicle parameter (re)calibration problem relying on the comparison of the vehicle's motion observed by two sensor sources. 
We assumed that the static objects observed in the environment move in accordance to the vehicle motion.
In this study the calibration of a vehicle's parameters was achieved by optimization.
This chapter introduces the explanation of the applied feature extraction followed by the formulation of the optimization problem as well as the definition of the kinematic models.
\subsection{Feature Extraction}
\begin{figure}[t]
	\centering
	\vspace{0.3cm}
	\includegraphics[width=0.48\textwidth]{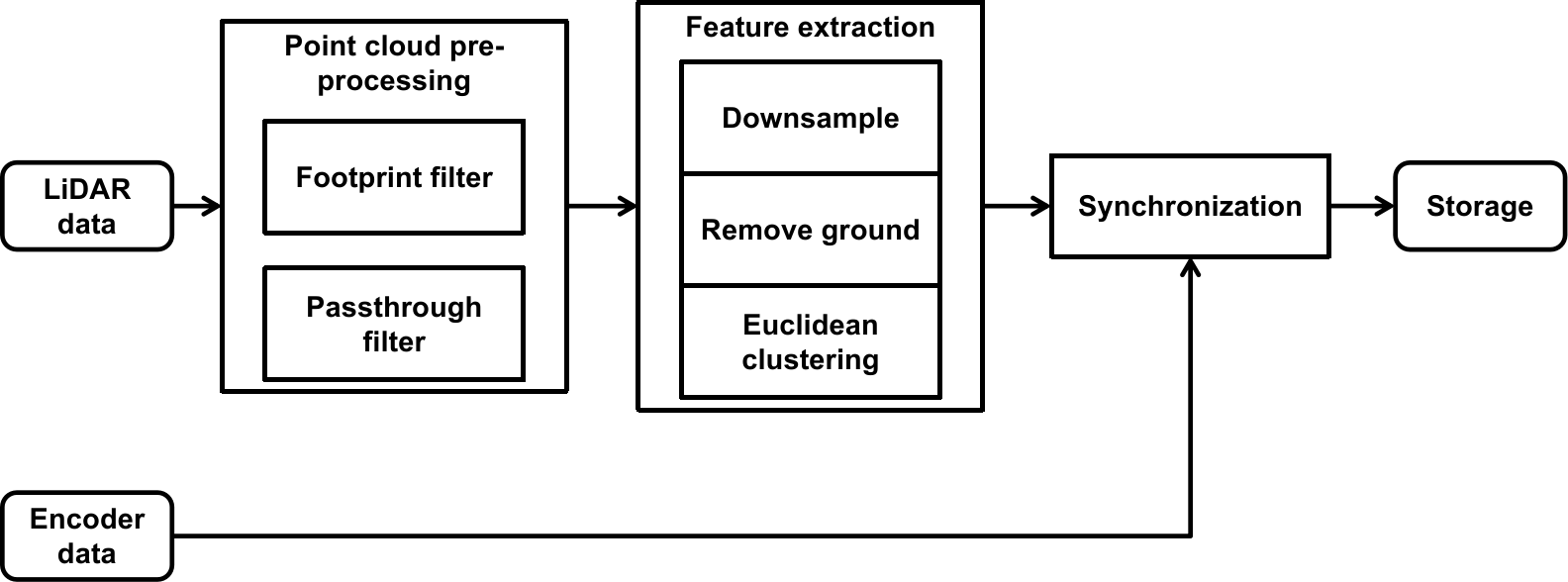}
	\caption{Feature extraction pipeline}
	\label{fig:FeatureExtraction}
\end{figure}

The following subsection explains the feature extraction approach utilized in our method, which is visualized in Fig. \ref{fig:FeatureExtraction}.
As depicted, the feature extraction consists of two major parts: point cloud pre-processing and feature extraction, which were both implemented using the pcl\_ros API\footnote{\url{https://github.com/ros-perception/perception_pcl/tree/melodic-devel/pcl_ros}}. In the first part, we utilize CropBox and PassThrough filters to remove any vehicle-related measurements and noise. After pre-processing we handled the feature extraction by downsampling the pre-processed point cloud, followed by a ground removal process using RANSAC Perpendicular Plane Segmentation, and finally, we applied euclidean clustering to estimate the center of the landmark. Finally, the cluster and the encoder data are synchronized and stored for our parameter estimation approach.
\subsection{Optimization for Vehicle Parameters}
To this end we formulated the optimization problem by observing the vehicle's motion from time step $t-1$ to $t$.
The motion was observed relying on wheel angular motion as well as the motion of the vehicle relative to the landmark.

In time step $t-1$ we measured the distance to the landmark $\vec{L}_{t-1}=\begin{pmatrix}x_{l_{t-1}}, y_{l_{t-1}}\end{pmatrix}^T$ from the vehicle's pose $\vec{x} = \left(x_{t-1},y_{t-1},\theta_{t-1}\right)^T$.
From time step $t-1$ to $t$, we further observed the same landmark at $\vec{L}_{t}=\begin{pmatrix}x_{l_{t}}, y_{l_{t}}\end{pmatrix}^T$. 
Furthermore, we stored actuator encoder data and used this information to formulate a motion equation $\vec{\rho}=f\left( \left<\Delta\vec{\phi}, \Delta\delta\right>,\Theta\right)$, 
where $\Delta\vec{\phi}$ describes the angular difference of the wheels from $t-1$ to $t$, $\Delta\delta$ describes the change of steering angle and $\Theta$ describes the set of vehicle parameters, such as wheel radius $r$, baseline $B$ and wheel base $L$.

\begin{figure}[!t]
	\centering
	\includegraphics[width=0.33\textwidth]{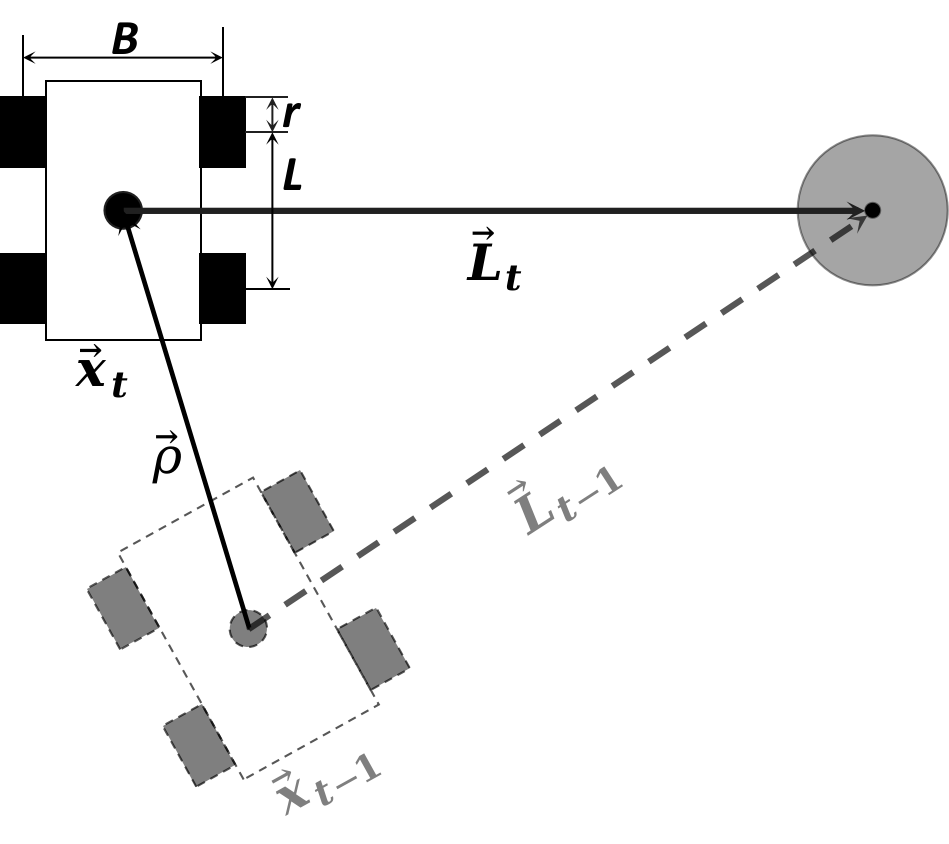}
	\caption{Proposed method from time step $t-1$ to $t$. With the robot's 2D pose $\vec{x}_{t-1}$ respectively $\vec{x}_t$, the landmark distances $\vec{L}_{t-1}$ and $\vec{L}_{t}$, and $\mid\mid\vec{\rho}\mid\mid$ the traversed distance in $\Delta t$. $r$, $B$, and $L$ are the desired wheel radii, track distance and wheel base respectively.}
	\label{fig:Method}
\end{figure}

We assumed smooth wheel motion and a rigid body, thus the true vehicle motion as well as the landmark motion must be in accordance.
This assumption is visualized in Fig. \ref{fig:Method}.

Relying on smooth wheel motion we optimized the vehicle parameters in order to minimize the difference between $\vec{\rho}$ and the vehicle motion calculated by landmark observation.
This optimization problem is formulated in \eqref{eq:OPT1}
\begin{align}\label{eq:OPT1}
    & \argmin{\Theta} \left(\sum_{i=0}^{n} \mid\mid \mathbf{R}_{i,t-1}^t\cdot \vec{L}_{i,t-1} - \vec{L}_t - \vec{\rho}_i \mid\mid_2^2  \right) 
\end{align}
where
\begin{align}
        \textbf{R}_{i,t-1}^t &=  \left(\begin{matrix} \cos{\theta_i} & -\sin{\theta_i}\\ \sin{\theta_i} & \cos{\theta_i} \\     \end{matrix}\right)_{t-1}^t \nonumber \\
    \vec{L}_{j} &= \left(\begin{matrix}{x_i}_{l_{j}}\\ {y_i}_{l_{j}}\end{matrix}\right) \nonumber \\
    \vec{\rho}_i &= \left(\begin{matrix} {\rho_i}_{x}\\ {\rho_i}_{y}\end{matrix}\right) \nonumber
\end{align}
and solved relying on a set of motion and sensor observations with sequential least squares programming (SLSQP) \cite{Optimizer} by minimizing the squared $l^2-norm$ of the loss function defined in \eqref{eq:OPT1}.

\subsection{Kinematic Configurations}
In order to quantify the performance of the proposed pipeline we used different kinematic configurations by relying on simulated data.

\subsubsection{Differential Drive Model}
\begin{figure}[t]
    \centering
    \includegraphics[width=0.35\textwidth]{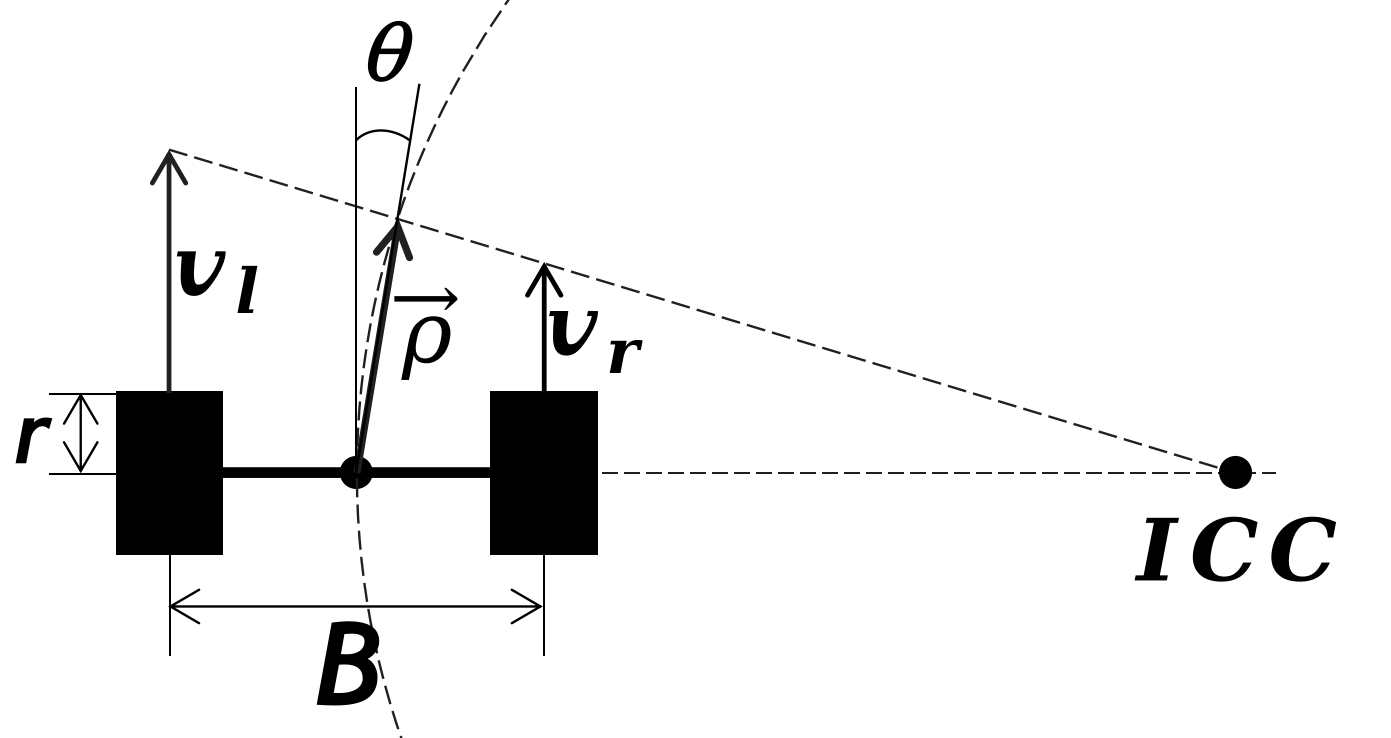}
    \caption{Differential drive kinematics. $v_{l,r}$ are the wheel velocities, $\theta$ is the change of orientation and $\mathbf{ICC}$ is the Instantaneous Center of Curvature. $r$ and $B$ are the wheel radius and baseline.}
    \label{fig:diff_drive}
\end{figure}

As depicted in Fig. \ref{fig:diff_drive} differential drive vehicles solely rely on the change of wheel speeds to steer the agent on the 2D plane. The distance traveled by the origin of the robot is thus only dependent on the change of wheel speeds as \eqref{eq:diff:rho} points out \cite{Siegwart:AMR}. The change of heading is conditioned on the change of wheel speeds and the baseline of the mobile agent \eqref{eq:diff:theta} \cite{Siegwart:AMR}.

\begin{align}
    \rho =& \frac{v_l + v_r}{2} \cdot \Delta t= \frac{\omega_l + \omega_r}{2}  \cdot r \cdot \Delta t = \frac{\Delta\phi_l + \Delta\phi_r}{2} \cdot r \label{eq:diff:rho}\\
    \theta =& \frac{\Delta\phi_r - \Delta\phi_l}{B} \cdot r \label{eq:diff:theta}
\end{align}

\subsubsection{Bicycle Model}
The bicycle model is a frequently used simplification of the profound Ackerman steering model \cite{Bicycle, KinBicycle}. The traveled distance $\rho$ between two time steps can be computed by considering the change of wheel speed \eqref{eq:bicycle:rho} whereas the change of heading can be computed as described in \eqref{eq:Bicycle:theta}. The combined steering angle of the bicycle model can be expressed using the actual inner ($\delta_i$) and outer  ($\delta_o$) steering angle of the Ackerman model as seen in \eqref{eq:Bicycle:delta}.
\begin{figure}[t]
    \centering
    \includegraphics[width=0.4\textwidth]{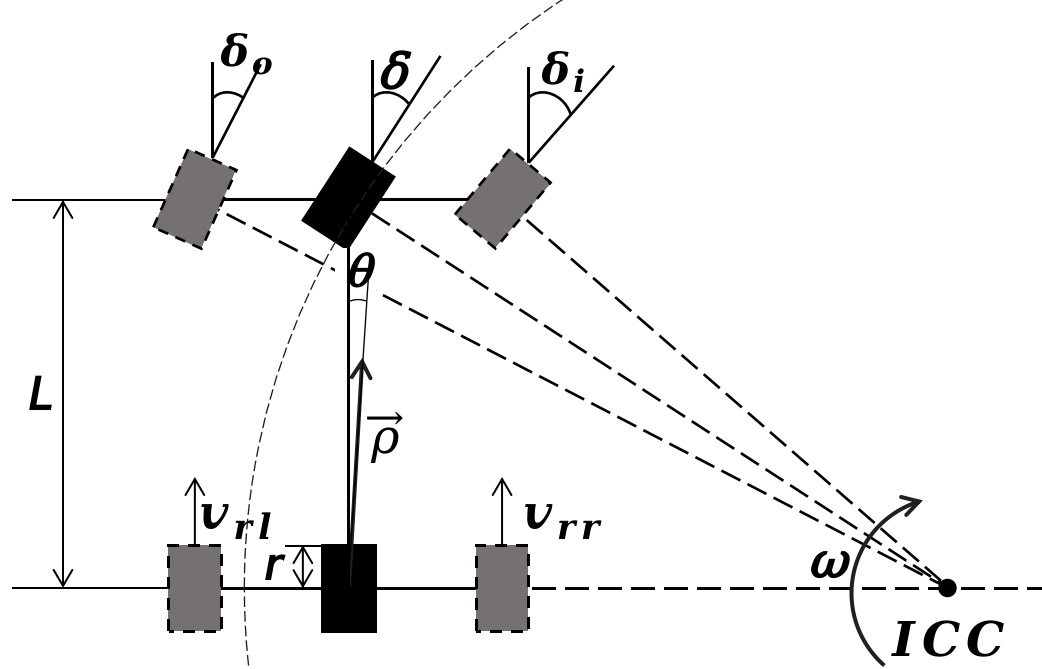}
    \caption{Bicycle kinematics. $v_{rl,rr}$ are the rear wheel velocities, and $\omega$ is the angular velocity around the $\mathbf{ICC}$. $r$ and $L$ are the wheel radius and wheel base.}
    \label{fig:bicycle}
\end{figure}

\begin{align}
    \rho =& v \cdot \Delta t = \omega_{wheels} \cdot r \cdot \Delta t = \Delta\phi \cdot r \cdot \frac{1}{\Delta t} \label{eq:bicycle:rho}\\
    \theta = & \theta_{t-1} + \omega \cdot \Delta t \label{eq:Bicycle:theta}\\
    \cot{\delta} =& \frac{\cot{\delta_o}}{\cot{\delta_i}} \label{eq:Bicycle:delta}
\end{align}

where

\begin{align}
    \Delta\phi =& \frac{\Delta\phi_{rl} + \Delta\phi_{rr}}{2} \nonumber\\
    \omega =& \frac{v \cdot \tan\left(\delta\right)}{L} \nonumber
\end{align}

\section{Experimental Setup}\label{sec:4}
\begin{figure}[h]
    \centering
    \begin{subfigure}[t]{0.27\textwidth}
        \includegraphics[width=\textwidth]{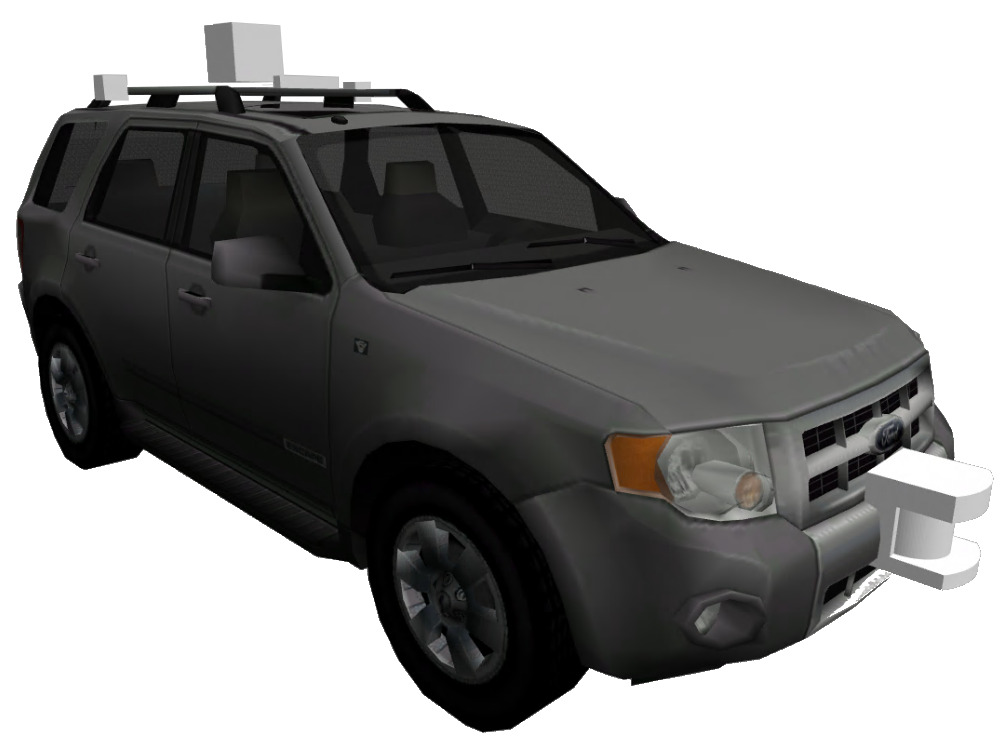}
        \caption{Modified Catvehicle \cite{Catvehicle} simulation with  360$^{\circ}$ 64 layer Velodyne LiDAR: Ackerman steering}
        \label{fig:Catvehicle}
    \end{subfigure}
    \hfill
    \begin{subfigure}[t]{0.16\textwidth}
        \includegraphics[width=\textwidth]{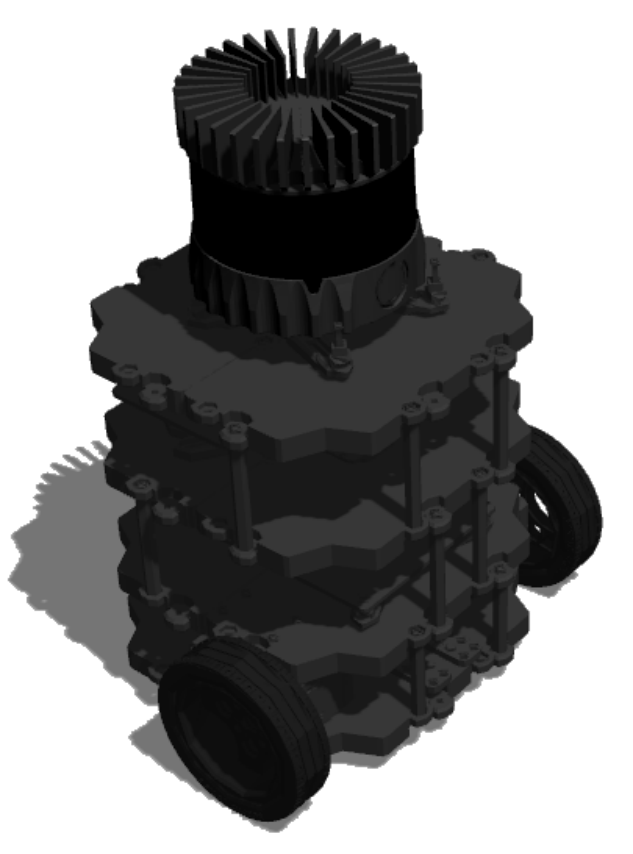}
        \caption{Turtlebot3 Burger simulation with 360$^{\circ}$ 64 layer Velodyne LiDAR, educational robot: differential drive}
        \label{fig:turtlebot}
    \end{subfigure}
            
    \caption{Mobile vehicles used in this study. In order to investigate the applicability of the proposed framework, we implemented two vehicle simulations (an autonomous car in Fig. \ref{fig:Catvehicle} and a mobile robot in Fig. \ref{fig:turtlebot}).
    The used vehicles rely on different kinematic configurations.}
    \label{fig:vehicles}
\end{figure}

This section explains the experimental setup and describes the mobile vehicles, as seen in Fig. \ref{fig:vehicles}, utilized to validate the proposed calibration method.
We recorded the datasets with a simulation of a mobile differential drive platform\footnote{\url{https://github.com/ROBOTIS-GIT/turtlebot3}} and a simulated Ackerman steering vehicle\footnote{\url{https://github.com/jmscslgroup/catvehicle}} based on the Cognitive and Autonomous Test Vehicle (Catvehicle) \cite{Catvehicle} as depicted in Fig. \ref{fig:vehicles}. 

To this end we adopted a kinematic bicycle model for the simulated Ackerman vehicle and a differential drive model for the Turtlebot3 respectively, Fig. \ref{fig:Method} illustrates a generalization of the developed approach. For both, the differential drive as well as the bicycle model we assume constant control inputs during the time intervals $\left[t-1, t\right]$ and further assume $\Delta t$ to be reasonably small.
For the simulated differential drive and the Ackerman vehicle, we utilized the Robot Operating System (ROS)\cite{ROS} and the Point Cloud Library (PCL) \cite{PCL} to capture the distance data in combination with the joint state data for the encoder states. The simulation of the two vehicles was implemented in GAZEBO \cite{Gazebo} which simulates physics using the Open Dynamics Engine (ODE) \cite{ODE}, where friction and damping coefficients, sensor noise, gravity, buoyancy, and other parameters of the Gazebo models can be tuned to approximate genuine real world behavior. 
For the purpose of this study, the vehicle models of the Turtlebot as well as for the Catvehicle were updated by an additional 360$^{\circ}$ 64 layer Light Detection and Ranging (LiDAR)\footnote{\url{https://bitbucket.org/DataspeedInc/velodyne_simulator}} for distance estimation to the landmark and with a ground truth (GT) positioning sensor\footnote{\url{http://docs.ros.org/en/electric/api/gazebo_plugins/html/group__GazeboRosP3D.html}} to acquire the GT trajectory. 
As landmarks we chose cylindrical objects with a diameter of 0.1m and a height of 0.4m for the differential drive simulation and 0.5m in diameter and 0.4m in height for the Ackerman simulation. In comparison to other objects such as cuboids, which may not be mapped completely due to the 2.5D shadow cast by the LiDAR sensor, we utilized cylindrical objects since they make it easier to determine the center.

\begin{figure}[h]
	\centering
	\begin{subfigure}[t]{0.225\textwidth}
			\vspace{0.2cm}\includegraphics[width=\textwidth]{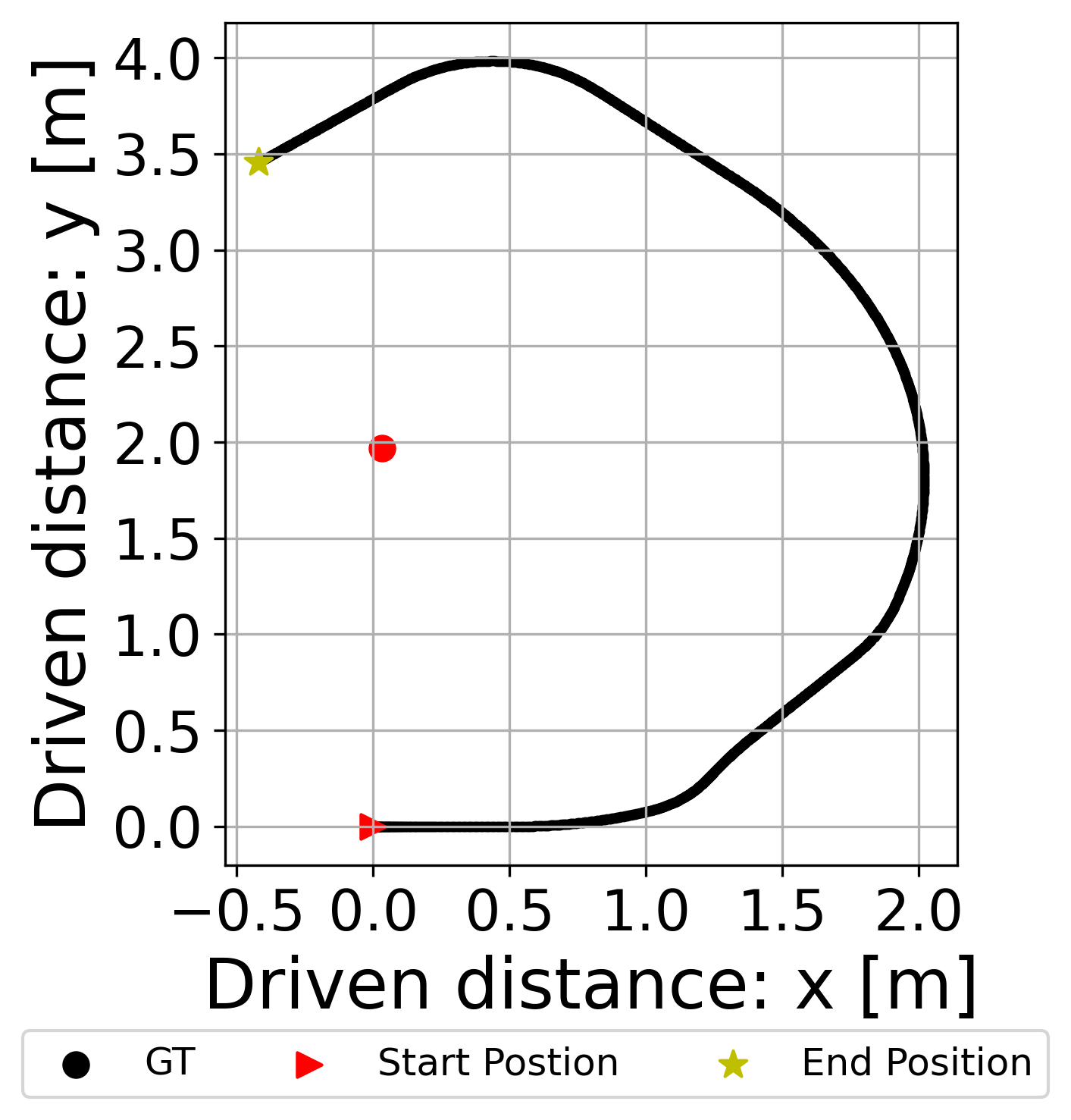}
		\caption{GT trajectory of the differential drive dataset, containing 108 straight and 293 turn datapoints} 
		\label{fig:Turtlebot:Gt}
	\end{subfigure}
	\hfill
	\begin{subfigure}[t]{0.24\textwidth}
		\vspace{0.2cm}\includegraphics[width=\textwidth]{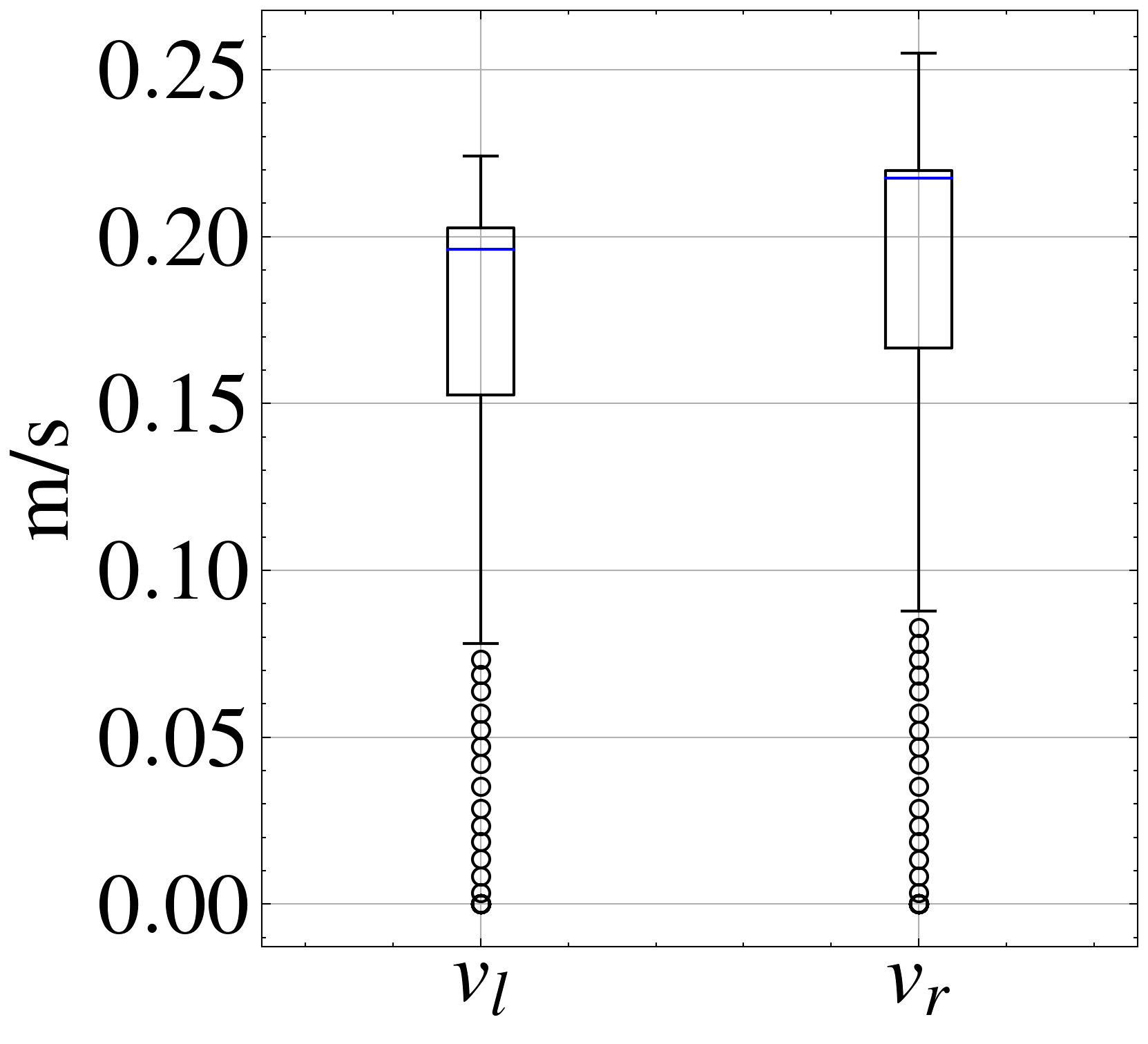}
		\caption{Velocities of the left ($v_l$) and right ($v_r$) wheels of the differential drive dataset} 
		\label{fig:Turtlebot:WheelSpeeds}
	\end{subfigure}
    \hfill
	\centering
	\begin{subfigure}[t]{0.225\textwidth}
		\vspace{0.2cm}\includegraphics[width=\textwidth]{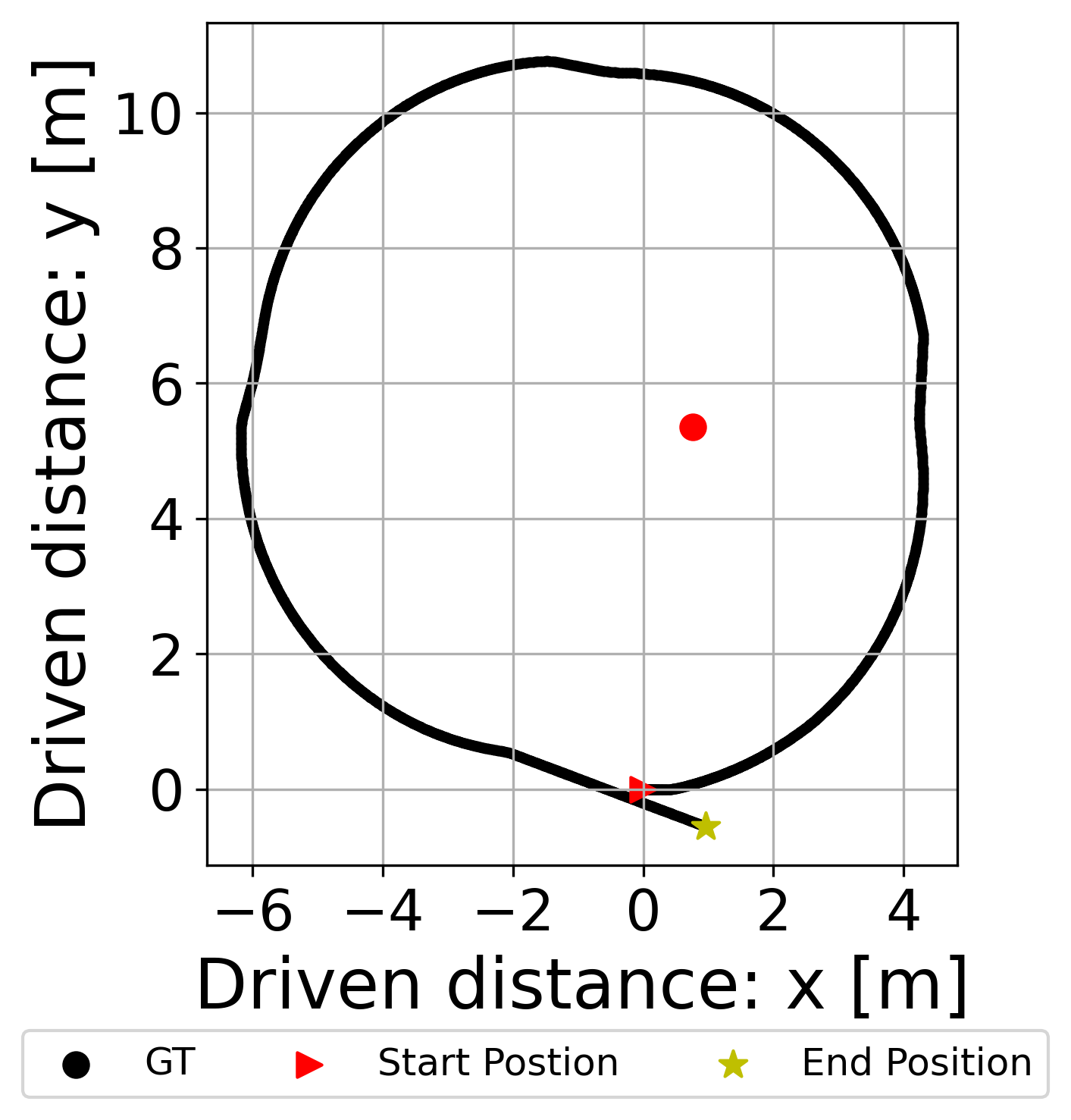}
		\caption{GT trajectory of the Catvehicle dataset, containing 77 straight and 824 turn datapoints} 
		\label{fig:Catvehicle:Gt}
	\end{subfigure}
	\hfill
	\begin{subfigure}[t]{0.24\textwidth}
		\vspace{0.2cm}\includegraphics[width=\textwidth]{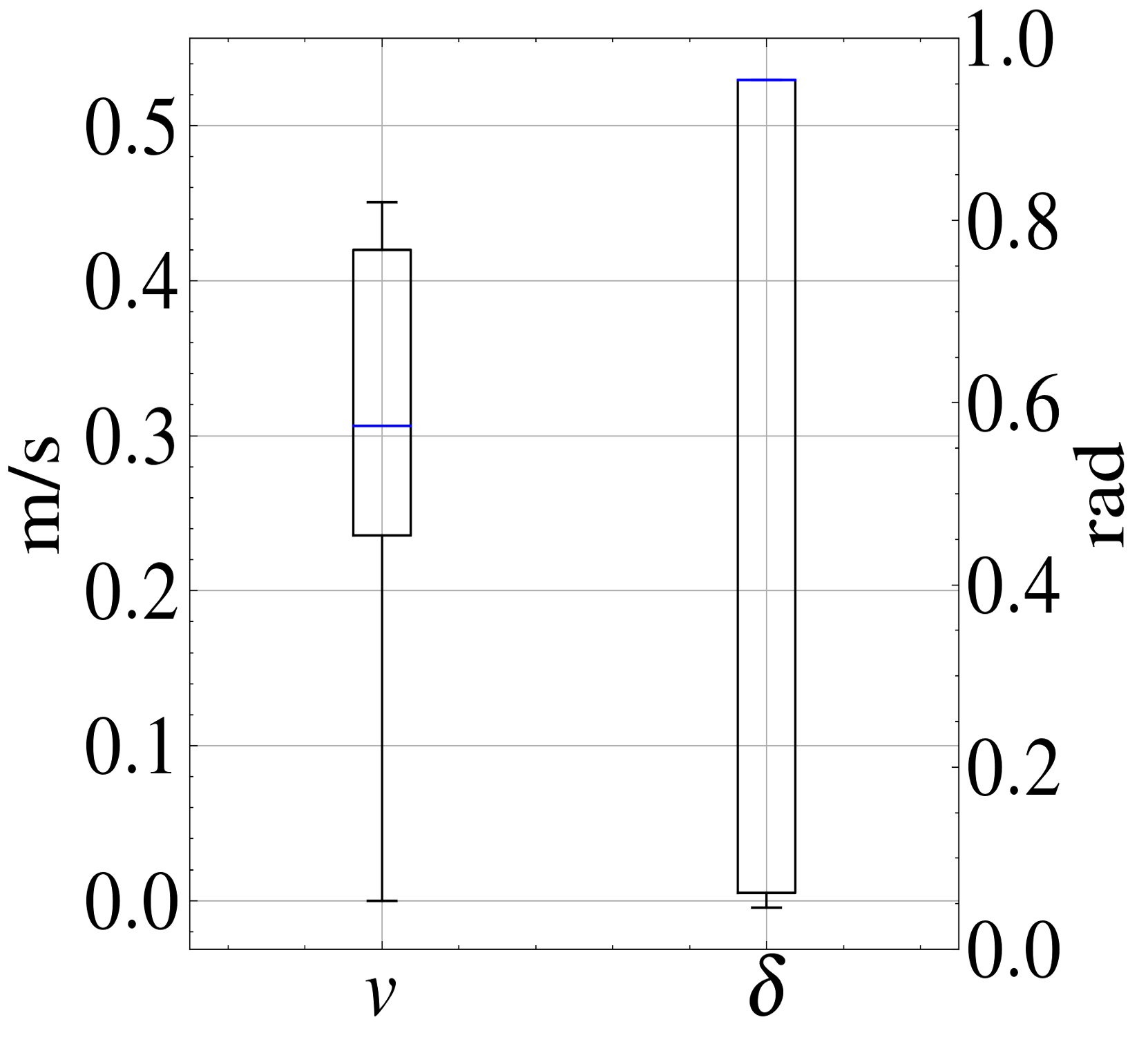}
		\caption{Velocity of the combined rear wheel ($v$) and the steering angle ($\delta$) of the Catvehicle dataset} 
		\label{fig:Catvehicle:WheelSpeeds}
	\end{subfigure}
	\caption{40 seconds long Turtlebot3 data set (top) and 90 seconds long Catvehicle data set (bottom) consisting of 401 encoder and cluster datapoints for the Turtlebot and 901 encoder and cluster data points for the Catvehicle.}\label{fig:Datasets}
\end{figure}

The Turtlebot3 is an educational differential drive mobile robot \cite{Turtlebot3Edu} produced by ROBOTIS \cite{TurtleBot:online}. 
Fig. \ref{fig:Turtlebot:Gt} displays the ground truth trajectory of the dataset as well as the corresponding wheel velocities (Fig. \ref{fig:Turtlebot:WheelSpeeds}), generated by the joint states of the left and right wheel $v_l$ and $v_r$ respectively.

For the Ackerman steering simulation we adapted the Catvehicle, a research test-bed for autonomous driving technology\cite{Catvehicle}, which is based on a Ford Hybrid Escape. The following Fig. \ref{fig:Catvehicle:Gt} displays the GT trajectory of the dataset, provided by the GT positioning sensor, as well as the corresponding wheel velocities $v$ (Fig. \ref{fig:Catvehicle:WheelSpeeds}), generated by the rear joint states of the left and right wheel and the combined steering angle $\delta$ respectively.

\begin{figure*}[t]
	\centering
	\vspace{0.2cm}
	\begin{subfigure}[t]{0.24\textwidth}
		\includegraphics[width=\textwidth]{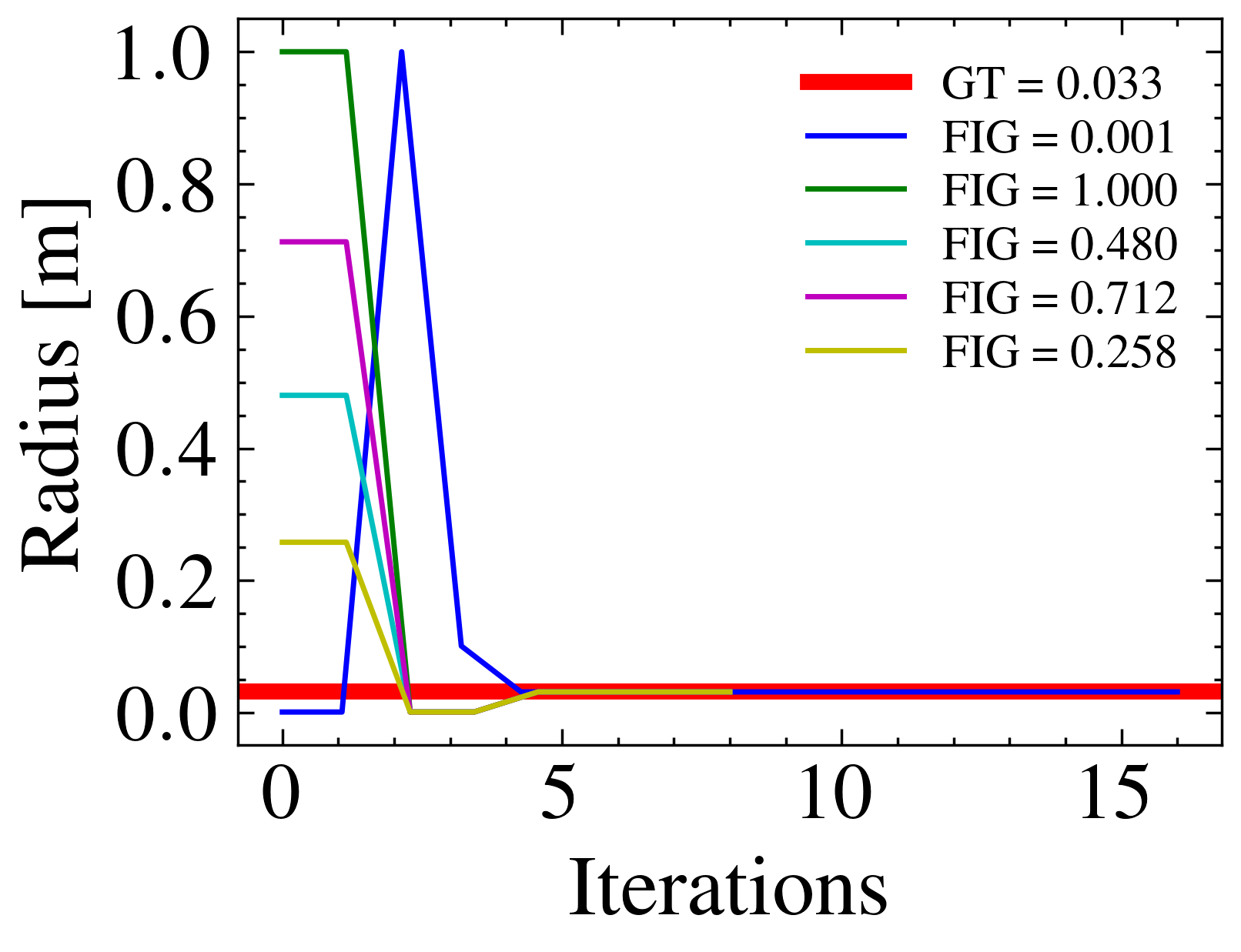}
		\caption{Turtlebot3 optimizations: Radius} 
		\label{fig:DiffOptim:Radius}
	\end{subfigure}
	\hfill
	\begin{subfigure}[t]{0.24\textwidth}
		\includegraphics[width=\textwidth]{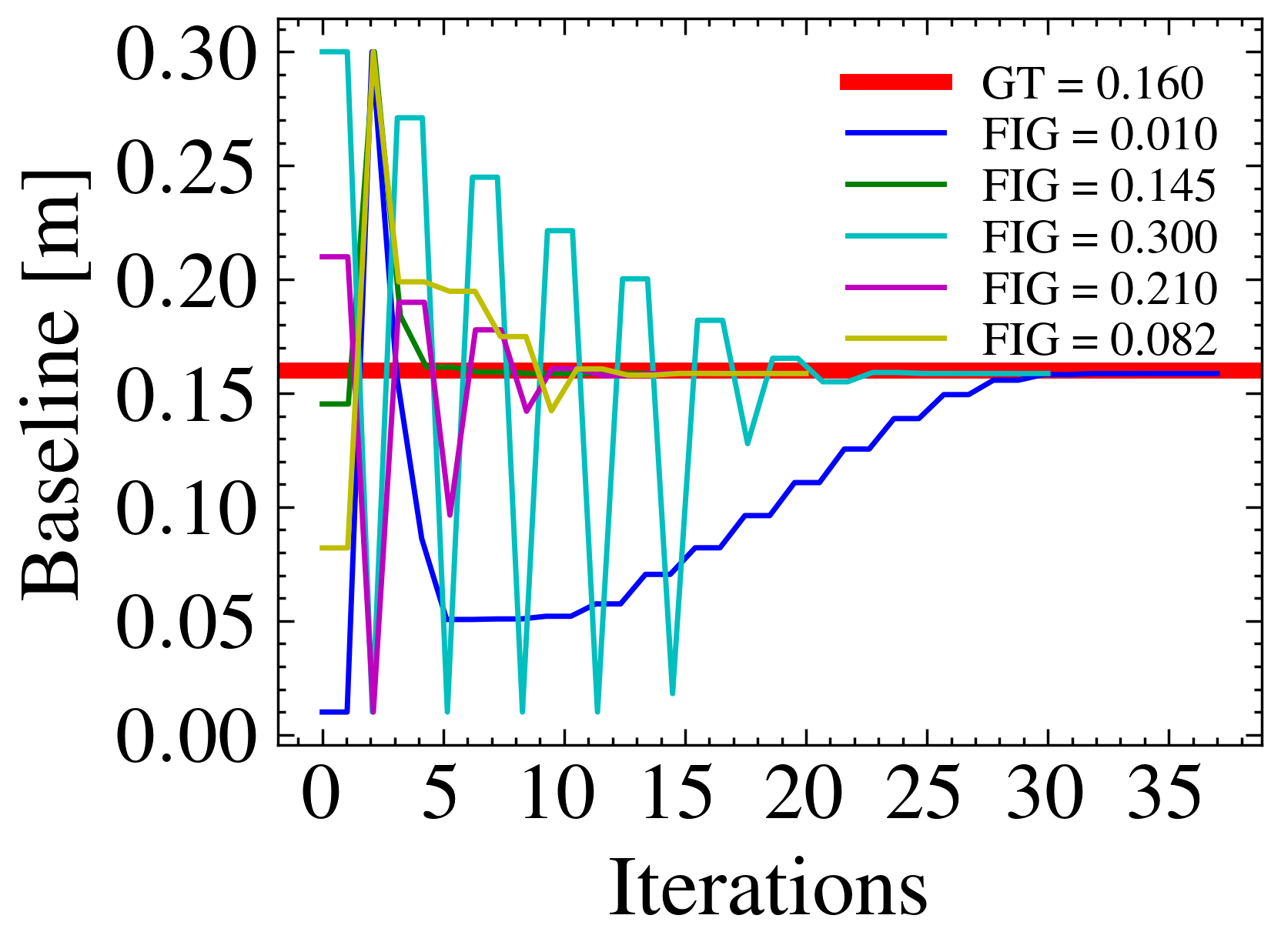}
		\caption{Turtlebot3 optimizations: Baseline} 
		\label{fig:DiffOptim:Baseline}
	\end{subfigure}
	\begin{subfigure}[t]{0.24\textwidth}
		\includegraphics[width=\textwidth]{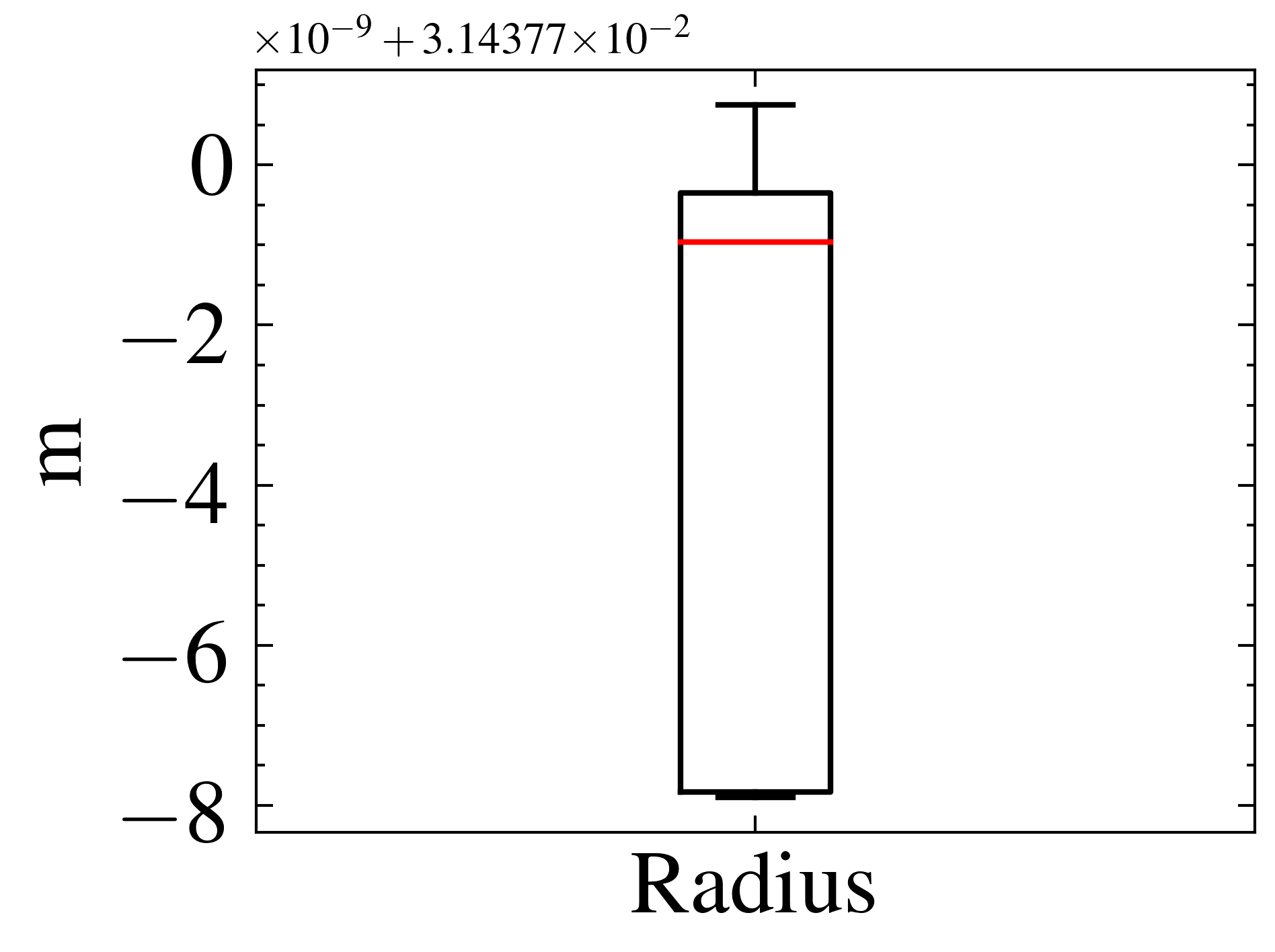}
		\caption{Turtlebot3 radius optimization results of 100 iterations} 
		\label{fig:DiffLoss:Radius}
	\end{subfigure}
	\hfill
	\begin{subfigure}[t]{0.24\textwidth}
		\includegraphics[width=\textwidth]{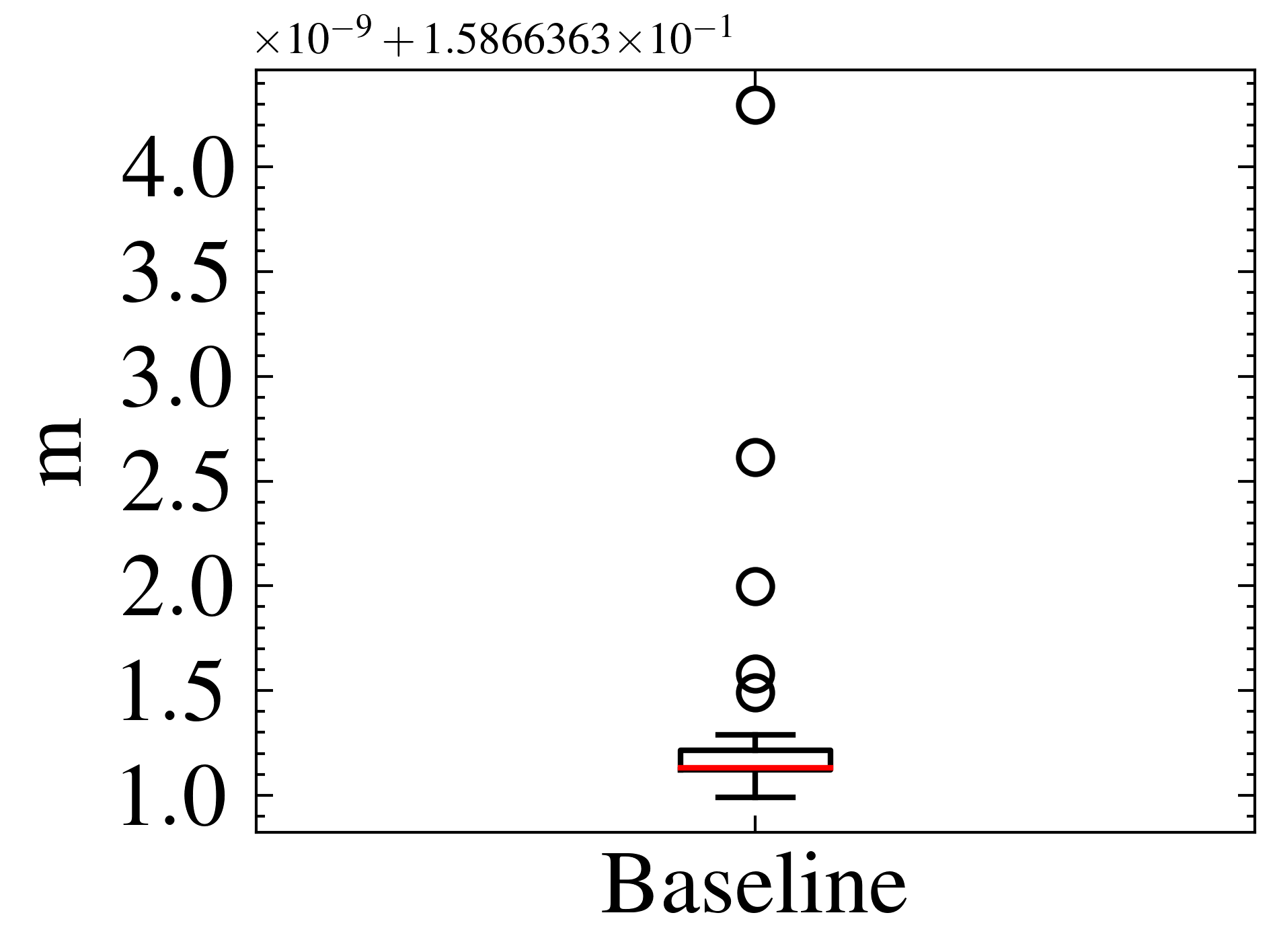}
		\caption{Turtlebot3 baseline optimization results of 100 iterations} 
		\label{fig:DiffLoss:Baseline}
	\end{subfigure}
	
	\vspace{0.3cm}
	\begin{subfigure}[t]{0.24\textwidth}
		\includegraphics[width=\textwidth]{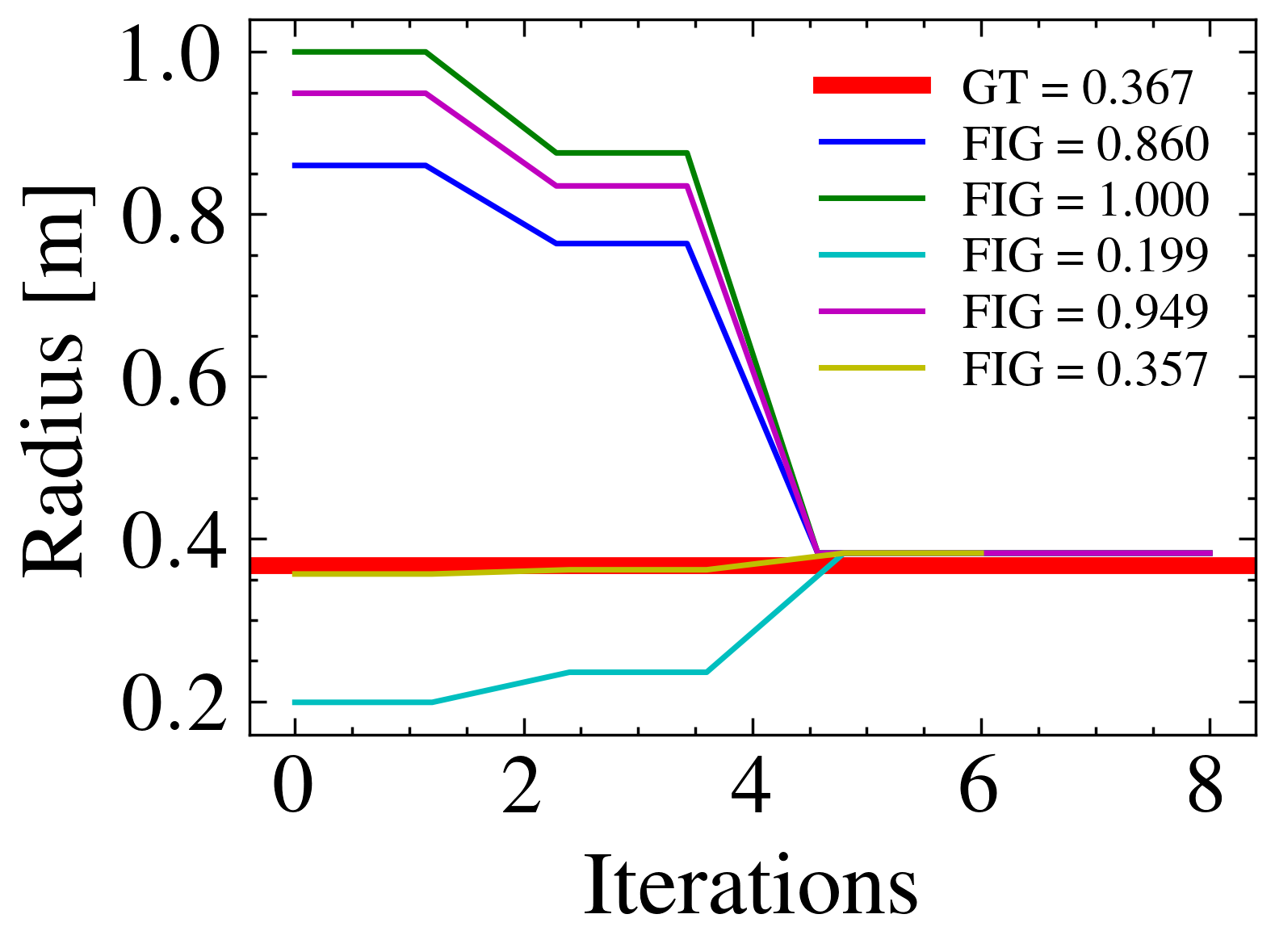}
		\caption{Catvehicle optimizations: Radius} 
		\label{fig:BicycleOptim:Radius}
	\end{subfigure}
	\hfill
	\begin{subfigure}[t]{0.24\textwidth}
		\includegraphics[width=\textwidth]{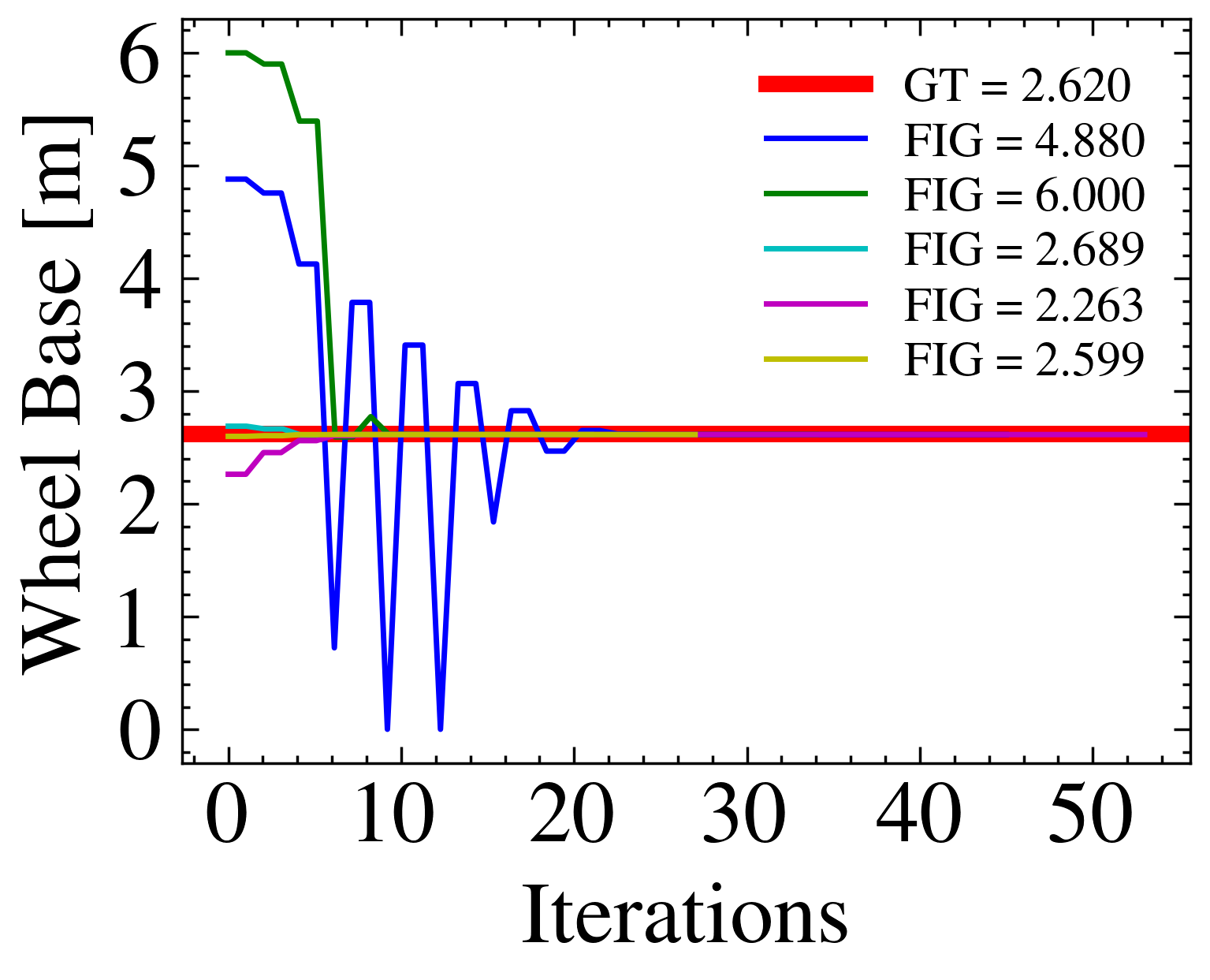}
		\caption{Catvehicle optimizations: Wheel base} 
		\label{fig:BicycleOptim:Baseline}
	\end{subfigure}
	\begin{subfigure}[t]{0.24\textwidth}
		\includegraphics[width=\textwidth]{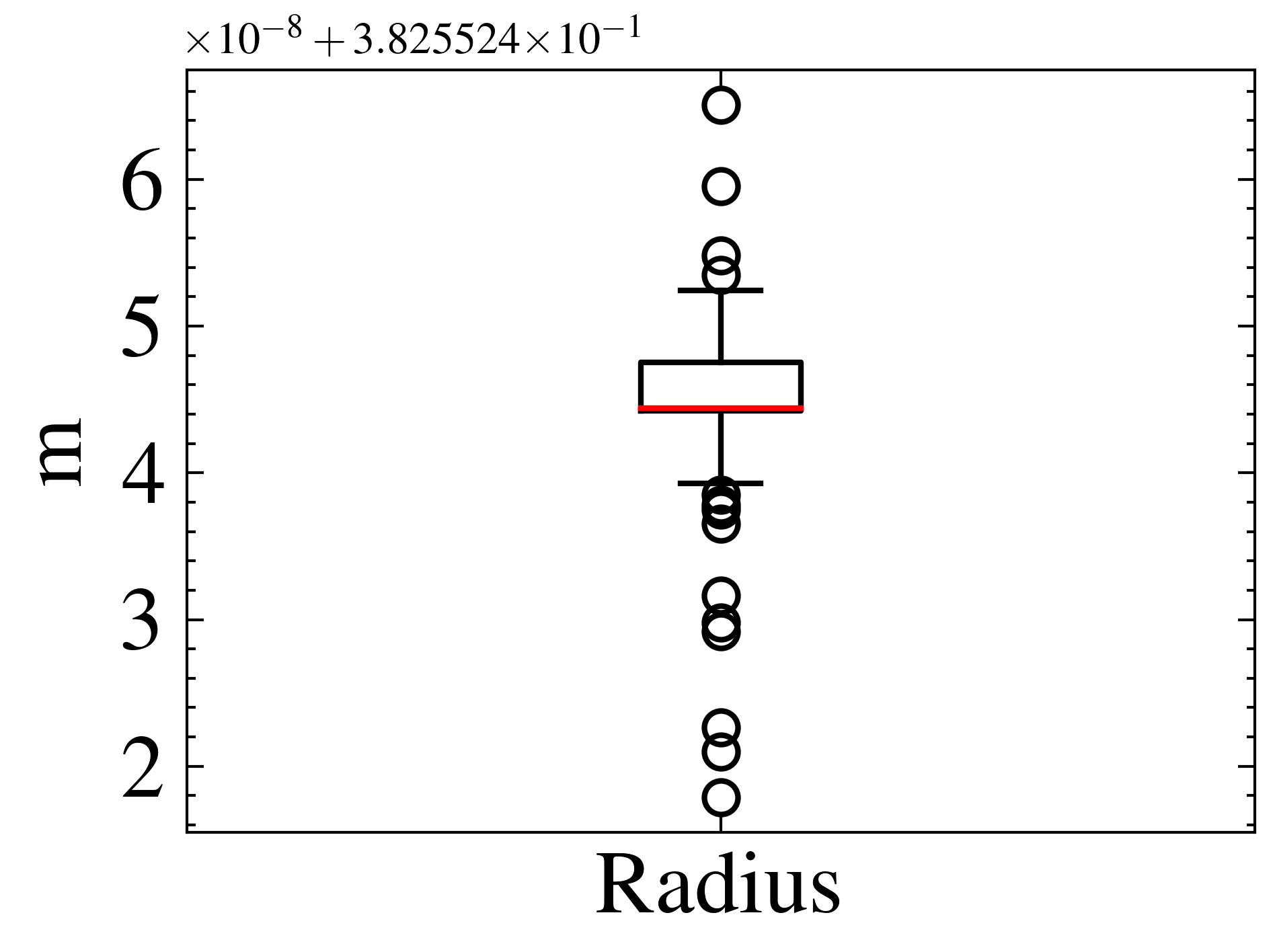}
		\caption{Catvehicle radius optimization results of 100 iterations} 
		\label{fig:BicycleLoss:Radius}
	\end{subfigure}
	\hfill
	\begin{subfigure}[t]{0.24\textwidth}
		\includegraphics[width=\textwidth]{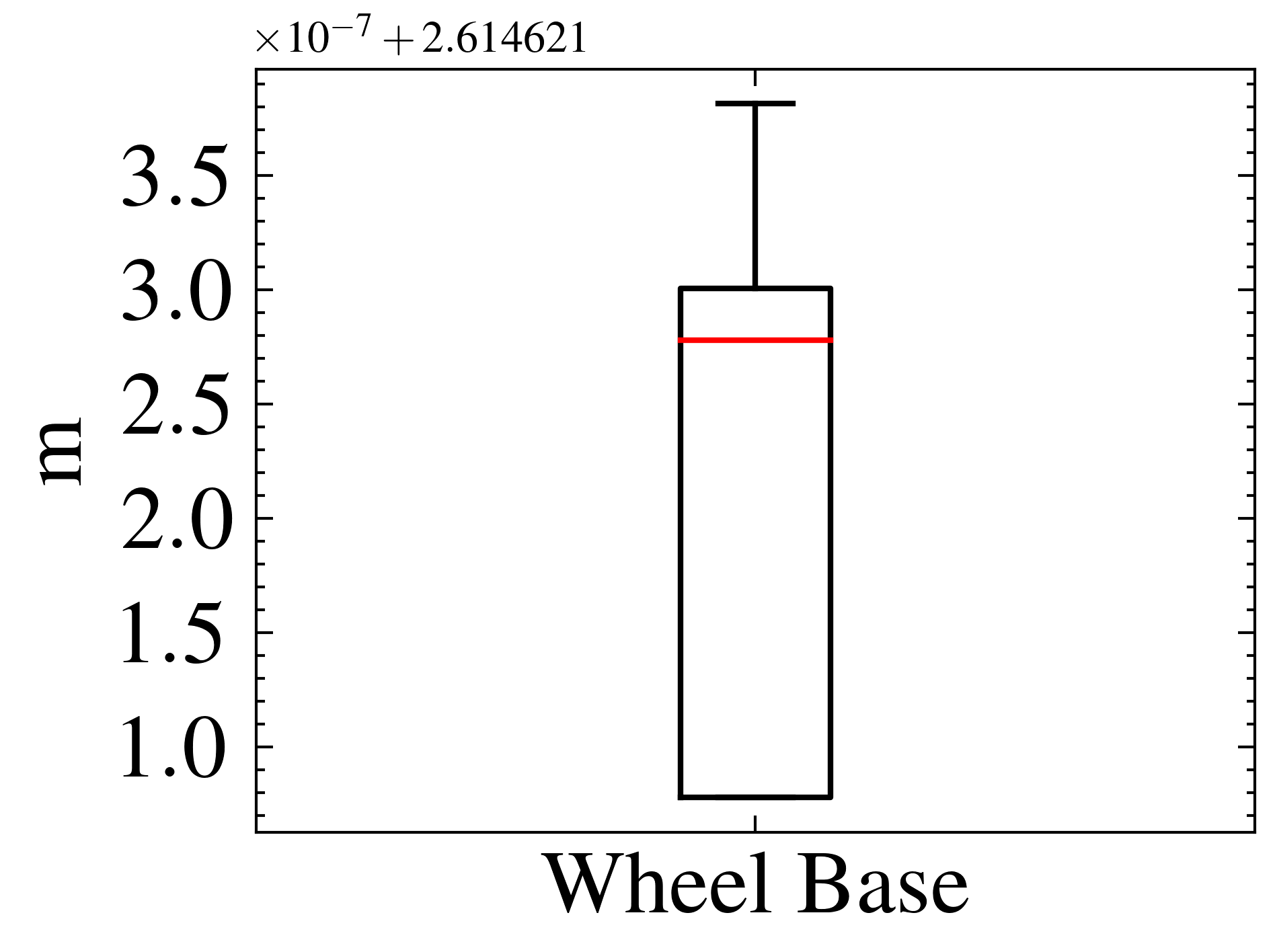}
		\caption{Catvehicle wheel base optimization results of 100 iterations} 
		\label{fig:BicycleLoss:Baseline}
	\end{subfigure}
	\caption{Optimization results of the simulated Turtlebot3 Burger (top) and the Catvehicle (bottom) compared with the GT values in red and the corresponding distributions.\label{fig:Results}}
\end{figure*}

In order to obtain reliable and correct results we split the optimization procedure in two steps, namely radius estimation on a straight part of the track and baseline optimization on curvy parts of the track. For the differential drive vehicle the straight/turn dataset were created by splitting in accordance to the difference of wheel speeds ($\Delta v = \Delta v_r - \Delta v_l$) whereas for the Ackerman steering the straight/turn dataset were created in accordance to the steering angle $\delta$.

\section{Results}\label{sec:5}
The following section presents and discusses the kinematic parameters obtained via the proposed optimization approach and further compares them to the GT values, taken from the corresponding Universal Robot Description Format (URDF) files. Table \ref{tab:Results} displays the resulting mean values of a total of 100 optimizations with different first initial guesses (FIG) for each parameter of the different kinematic configurations as well as the error with respect to the ground truth values. To obtain the FIG values we sampled from a normal distribution with the mean located at the GT value. 

\begin{table}[b]
    \caption{Results of 100 optimizations}
    \label{tab:Results}
    \begin{tabular}{|l|ll|ll|ll|}
        \hline
        \multicolumn{1}{|c|}{\multirow{2}{*}{\textbf{Vehicle}}} & \multicolumn{2}{c|}{\textbf{GT [m]}}& \multicolumn{2}{c|}{\textbf{Optimization [m]}}& \multicolumn{2}{c|}{\textbf{Error [\%]}}\\ \cline{2-7} 
        \multicolumn{1}{|c|}{}& \multicolumn{1}{c|}{r} & \multicolumn{1}{c|}{B/L} & \multicolumn{1}{c|}{r} & \multicolumn{1}{c|}{B/L} & \multicolumn{1}{c|}{r} & \multicolumn{1}{c|}{B/L} \\ \hline
        Turtlebot3& \multicolumn{1}{l|}{0.033}& 0.16& \multicolumn{1}{l|}{0.03144}& 0.15866& \multicolumn{1}{l|}{4.73}& 0.84\\ \hline
        Catvehicle& \multicolumn{1}{l|}{0.03672}& 2.62& \multicolumn{1}{l|}{0.38255}& 2.61462& \multicolumn{1}{l|}{4.18}& 0.21\\ \hline
    \end{tabular}
\end{table}

Fig. \ref{fig:Results} displays 5 SLSQP optimizations processes for each parameter of the kinematic models, based on the loss functions as defined in \eqref{eq:OPT1}, and the corresponding boxplots. 
As illustrated in Fig. \ref{fig:Turtlebot:Gt} and \ref{fig:Catvehicle:Gt} the curved path segments outweigh the straight path segments in the recorded datasets, this explains the more precise optimization of the baseline (Fig. \ref{fig:DiffOptim:Baseline}) and wheel base (Fig. \ref{fig:BicycleOptim:Baseline}) of the Turtlebot3 respectively the Catvehicle. 
Meanwhile, because there are fewer datapoints available in the straight path segments, the wheel radius optimization for the Turtlebot and the Catvehicle converges after 4 iterations (Fig. \ref{fig:DiffOptim:Radius}) or 5 iterations (Fig. \ref{fig:DiffOptim:Radius}), and the baseline (14 iterations, see Fig. \ref{fig:DiffOptim:Baseline}) or wheelbase (6 iterations, see Fig. \ref{fig:BicycleOptim:Baseline}) took longer to achieve the desired optimization tolerance of $1e^{-16}$.

The here proposed pipeline takes less sensor measurements (60 sec on average) compared to other approaches (e.g. 15 min in \cite{galasso2019efficient}) to produce reasonably precise estimations of the kinematic parameters. Further the presented method works with an arbitrary trajectory which makes it easier to carry out compared to approaches such as \cite{LEE2010582} that require a programmed trajectory for calibration. Meanwhile, as can be seen in table \ref{tab:ResultTime}, the optimization problem can be solved fairly fast.

\begin{table}[b]
    \centering
    \caption{Mean optimization duration of 100 optimizations on AMD Ryzen 7 3700X}
    \label{tab:ResultTime}
    \begin{tabular}{|c|cl|cl|}
        \hline
        \multirow{2}{*}{\textbf{Vehicle}} & \multicolumn{2}{c|}{\textbf{Turtlebot}} & \multicolumn{2}{c|}{\textbf{Catvehicle}} \\ \cline{2-5} 
         & \multicolumn{1}{c|}{r} & \multicolumn{1}{c|}{B} & \multicolumn{1}{c|}{r} & \multicolumn{1}{c|}{L} \\ \hline
        \multicolumn{1}{|l|}{\textbf{Time [s]}} & \multicolumn{1}{l|}{0.0397} & 0.2206 & \multicolumn{1}{l|}{0.0220} &0.8061  \\ \hline
    \end{tabular}
\end{table}

\section{Conclusion and Future Steps}\label{sec:6}
In this paper, we proposed a novel pipeline for parameter estimation of kinematic models. The pipeline applied minimizes the error of the relative position displacements of an arbitrary landmark compared to the forward kinematics of the vehicle by optimizing the desired parameters using sequential least square optimization. The error, compared to the GT values, of the estimated parameters range between $\approx$ 4.7\% with 40s of data and $\approx$ 0.2\% with 90s of data. 

Since the proposed pipeline heavily depends on the tracking of the observed landmark, occlusions will falsify the results. Therefore, future work will address the fusion of multiple landmarks, to overcome possible occlusions. Further, the development of an online capable pipeline as well as multiple radii estimations, to detect load changes and different abrasions will be tackled in future work.

\section*{ACKNOWLEDGMENT}
This work was supported by the Austrian Ministry for Climate
Action, Environment, Energy, Mobility, Innovation and
Technology (BMK) Endowed Professorship for Sustainable
Transport Logistics 4.0., IAV France S.A.S.U., IAV GmbH,
Austrian Post AG and the UAS Technikum Wien.

\bibliographystyle{IEEEtran}
\bibliography{bib}

\begin{thebibliography}{10}
\providecommand{\url}[1]{#1}
\csname url@rmstyle\endcsname
\providecommand{\newblock}{\relax}
\providecommand{\bibinfo}[2]{#2}
\providecommand\BIBentrySTDinterwordspacing{\spaceskip=0pt\relax}
\providecommand\BIBentryALTinterwordstretchfactor{4}
\providecommand\BIBentryALTinterwordspacing{\spaceskip=\fontdimen2\font plus
\BIBentryALTinterwordstretchfactor\fontdimen3\font minus
  \fontdimen4\font\relax}
\providecommand\BIBforeignlanguage[2]{{%
\expandafter\ifx\csname l@#1\endcsname\relax
\typeout{** WARNING: IEEEtran.bst: No hyphenation pattern has been}%
\typeout{** loaded for the language `#1'. Using the pattern for}%
\typeout{** the default language instead.}%
\else
\language=\csname l@#1\endcsname
\fi
#2}}

\bibitem{olaverri2017road}
C.~Olaverri-Monreal, ``Road safety: Human factors aspects of intelligent
  vehicle technologies,'' in \emph{Smart Cities, Green Technologies, and
  Intelligent Transport Systems}.\hskip 1em plus 0.5em minus 0.4em\relax
  Springer, 2017, pp. 318--332.

\bibitem{Siegwart:AMR}
R.~Siegwart and I.~R. Nourbakhsh, \emph{Introduction to Autonomous Mobile
  Robots}.\hskip 1em plus 0.5em minus 0.4em\relax USA: Bradford Company, 2004.

\bibitem{OdomNav}
S.~A.~S. Mohamed, M.-H. Haghbayan, T.~Westerlund, J.~Heikkonen, H.~Tenhunen,
  and J.~Plosila, ``A survey on odometry for autonomous navigation systems,''
  \emph{IEEE Access}, vol.~7, pp. 97\,466--97\,486, 2019.

\bibitem{OdomSlam1}
Z.~Xuexi, L.~Guokun, F.~Genping, X.~Dongliang, and L.~Shiliu, ``Slam algorithm
  analysis of mobile robot based on lidar,'' in \emph{2019 Chinese Control
  Conference (CCC)}, 2019, pp. 4739--4745.

\bibitem{OdomSlam2}
F.~Ducho{\v{n}}, J.~Ha{\v{z}}{\'\i}k, J.~Rodina, M.~T{\"o}lgyessy, M.~Dekan,
  and A.~Sojka, ``Verification of slam methods implemented in ros,''
  \emph{Journal of Multidisciplinary Engineering Science and Technology
  (JMEST)}, 2019.

\bibitem{Obstacle1}
C.~H\"ane, T.~Sattler, and M.~Pollefeys, ``Obstacle detection for self-driving
  cars using only monocular cameras and wheel odometry,'' in \emph{2015
  IEEE/RSJ International Conference on Intelligent Robots and Systems (IROS)},
  2015, pp. 5101--5108.

\bibitem{Obstacle2}
J.~Jin and W.~Chung, ``{Obstacle Avoidance of Two-Wheel Differential Robots
  Considering the Uncertainty of Robot Motion on the Basis of Encoder Odometry
  Information},'' \emph{Sensors}, vol.~19, no.~2, 2019.

\bibitem{8718624}
W.~Ci, Y.~Huang, and X.~Hu, ``Stereo visual odometry based on motion decoupling
  and special feature screening for navigation of autonomous vehicles,''
  \emph{IEEE Sensors Journal}, vol.~19, no.~18, pp. 8047--8056, 2019.

\bibitem{Thrun:PR}
S.~Thrun, W.~Burgard, and D.~Fox, \emph{Probabilistic Robotics (Intelligent
  Robotics and Autonomous Agents)}.\hskip 1em plus 0.5em minus 0.4em\relax The
  MIT Press, 2005.

\bibitem{Wober2020AutonomousKinematics}
W.~W{\"{o}}ber, G.~Novotny, L.~Mehnen, and C.~Olaverri-Monreal, ``{Autonomous
  vehicles: Vehicle parameter estimation using variational bayes and
  kinematics},'' \emph{Applied Sciences (Switzerland)}, vol.~10, no.~18, sep
  2020.

\bibitem{Ko:GPPF}
J.~Ko and D.~Fox, ``Gp-bayesfilters: Bayesian filtering using gaussian process
  prediction and observation models,'' in \emph{IROS}, 2008.

\bibitem{Samek:XAI1}
W.~Samek, T.~Wiegand, and K.-R. M\"uller, ``Explainable artificial
  intelligence: Understanding, visualizing, and interpreting deep learning
  models,'' \emph{ITU Journal: ICT Discoveries}, vol.~1, pp. 49--58, 2018.

\bibitem{Samek:XAI3}
W.~Samek and K.-R. M\"uller, \emph{Towards Explainable Artificial
  Intelligence}.\hskip 1em plus 0.5em minus 0.4em\relax Springer International
  Publishing, 2019, pp. 5--22.

\bibitem{Montavon:XAI1}
G.~Montavon, W.~Samek, and K.-R. M\"uller, ``Methods for interpreting and
  understanding deep neural networks,'' \emph{Digital Signal Processing},
  vol.~73, pp. 1--15, 2018.

\bibitem{KinBicycle}
P.~Polack, F.~Altché, B.~d'Andréa Novel, and A.~de~La~Fortelle, ``The
  kinematic bicycle model: A consistent model for planning feasible
  trajectories for autonomous vehicles?'' in \emph{2017 IEEE Intelligent
  Vehicles Symposium (IV)}, 2017, pp. 812--818.

\bibitem{lee2014accurate}
K.~Lee, C.~Jung, and W.~Chung, ``{Accurate calibration of kinematic parameters
  for two wheel differential mobile robots},'' \emph{Journal of Mechanical
  Science and Technology}, vol.~25, no.~6, pp. 1603--1611, jun 2011.

\bibitem{Sousa2020}
R.~B. Sousa, M.~R. Petry, and A.~P. Moreira, ``{Evolution of odometry
  calibration methods for ground mobile robots},'' in \emph{2020 IEEE
  International Conference on Autonomous Robot Systems and Competitions, ICARSC
  2020}.\hskip 1em plus 0.5em minus 0.4em\relax IEEE, apr 2020, pp. 294--299.

\bibitem{1512356}
G.~Antonelli, S.~Chiaverini, and G.~Fusco, ``{A calibration method for odometry
  of mobile robots based on the least-squares technique: theory and
  experimental validation},'' \emph{IEEE Transactions on Robotics}, vol.~21,
  no.~5, pp. 994--1004, oct 2005.

\bibitem{UMBmark}
\BIBentryALTinterwordspacing
J.~Borenstein and L.~Feng, ``{UMBmark: a benchmark test for measuring odometry
  errors in mobile robots},'' in \emph{Mobile Robots X}, W.~J. Wolfe and C.~H.
  Kenyon, Eds., vol. 2591, International Society for Optics and
  Photonics.\hskip 1em plus 0.5em minus 0.4em\relax SPIE, 1995, pp. 113--124.
  [Online]. Available: \url{https://doi.org/10.1117/12.228968}
\BIBentrySTDinterwordspacing

\bibitem{antonelli2007deterministic}
G.~Antonelli and S.~Chiaverini, ``{A Deterministic Filter for Simultaneous
  Localization and Odometry Calibration of Differential-Drive Mobile Robots},''
  in \emph{Proceedings of the 3rd European Conference on Mobile Robots - ECMR
  2007}, 2007, pp. 1--6.

\bibitem{DLTracking}
B.~Naujoks, P.~Burger, and H.-J. Wuensche, ``Combining deep learning and
  model-based methods for robust real-time semantic landmark detection,'' in
  \emph{2019 22th International Conference on Information Fusion (FUSION)},
  2019, pp. 1--8.

\bibitem{Optimizer}
D.~Kraft \emph{et~al.}, ``A software package for sequential quadratic
  programming,'' 1988.

\bibitem{galasso2019efficient}
F.~Galasso, D.~L. Rizzini, F.~Oleari, and S.~Caselli, ``Efficient calibration
  of four wheel industrial agvs,'' \emph{Robotics and Computer-Integrated
  Manufacturing}, vol.~57, pp. 116--128, 2019.

\bibitem{LEE2010582}
K.~Lee, W.~Chung, and K.~Yoo, ``{Kinematic parameter calibration of a car-like
  mobile robot to improve odometry accuracy},'' \emph{Mechatronics}, vol.~20,
  no.~5, pp. 582--595, 2010.

\bibitem{robotics5040023}
L.~Cantelli, S.~Ligama, G.~Muscato, and D.~Spina, ``{Auto-calibration methods
  of kinematic parameters and magnetometer offset for the localization of a
  tracked mobile robot},'' \emph{Robotics}, vol.~5, no.~4, 2016.

\bibitem{Martinelli2003}
A.~Martinelli, N.~Tomatis, A.~Tapus, and R.~Siegwart, ``Simultaneous
  localization and odometry calibration for mobile robot,'' in
  \emph{Proceedings 2003 IEEE/RSJ International Conference on Intelligent
  Robots and Systems (IROS 2003) (Cat. No.03CH37453)}, vol.~2, 2003, pp.
  1499--1504 vol.2.

\bibitem{kummerle2012simultaneous}
R.~K{\"{u}}mmerle, G.~Grisetti, and W.~Burgard, ``{Simultaneous Parameter
  Calibration, Localization, and Mapping},'' \emph{Advanced Robotics}, vol.~26,
  no.~17, pp. 2021--2041, dec 2012.

\bibitem{Deray2019JointManifold}
J.~Deray, J.~Sola, and J.~Andrade-Cetto, ``{Joint on-manifold self-calibration
  of odometry model and sensor extrinsics using pre-integration},'' in
  \emph{2019 European Conference on Mobile Robots (ECMR)}.\hskip 1em plus 0.5em
  minus 0.4em\relax IEEE, sep 2019, pp. 1--6.

\bibitem{Bicycle}
\BIBentryALTinterwordspacing
P.~Corke, ``Robotics, vision and control,'' vol. 118, 2017. [Online].
  Available: \url{http://link.springer.com/10.1007/978-3-319-54413-7}
\BIBentrySTDinterwordspacing

\bibitem{Catvehicle}
R.~Bhadani, J.~Sprinkle, and M.~Bunting, ``{The CAT Vehicle Testbed: A
  Simulator with Hardware in the Loop for Autonomous Vehicle Applications},''
  \emph{{Proceedings of 2nd International Workshop on Safe Control of
  Autonomous Vehicles (SCAV 2018), Porto, Portugal, 10th April 2018, Electronic
  Proceedings in Theoretical Computer Science 269, pp. 32–47}}, 2018.

\bibitem{ROS}
M.~Quigley, K.~Conley, B.~Gerkey, J.~Faust, T.~Foote, J.~Leibs, R.~Wheeler,
  A.~Y. Ng, \emph{et~al.}, ``Ros: an open-source robot operating system,'' in
  \emph{ICRA workshop on open source software}, vol.~3, no. 3.2.\hskip 1em plus
  0.5em minus 0.4em\relax Kobe, Japan, 2009, p.~5.

\bibitem{PCL}
R.~B. Rusu and S.~Cousins, ``3d is here: Point cloud library (pcl),'' in
  \emph{2011 IEEE international conference on robotics and automation}.\hskip
  1em plus 0.5em minus 0.4em\relax IEEE, 2011, pp. 1--4.

\bibitem{Gazebo}
N.~Koenig and A.~Howard, ``Design and use paradigms for gazebo, an open-source
  multi-robot simulator,'' in \emph{2004 IEEE/RSJ International Conference on
  Intelligent Robots and Systems (IROS) (IEEE Cat. No.04CH37566)}, vol.~3,
  2004, pp. 2149--2154 vol.3.

\bibitem{ODE}
R.~Smith \emph{et~al.}, ``Open dynamics engine,'' 2005.

\bibitem{Turtlebot3Edu}
R.~Amsters and P.~Slaets, ``Turtlebot 3 as a robotics education platform,'' in
  \emph{Robotics in Education}, M.~Merdan, W.~Lepuschitz, G.~Koppensteiner,
  R.~Balogh, and D.~Obdr{\v{z}}{\'a}lek, Eds.\hskip 1em plus 0.5em minus
  0.4em\relax Cham: Springer International Publishing, 2020, pp. 170--181.

\bibitem{TurtleBot:online}
ROBOTIS. Turtlebot 3.
  \url{https://emanual.robotis.com/docs/en/platform/turtlebot3/overview/}.
  (Accessed on 01/16/2022).

\end{thebibliography}
\end{document}